\documentclass[prx,twocolumn,superscriptaddress,floatfix,nopacs,nofootinbib]{revtex4-2}

\usepackage{graphicx,amsfonts,amssymb,amsmath,hyperref,enumerate}

\usepackage{algorithm}
\usepackage{algpseudocode}
\usepackage{pgfplots}
\usepackage{pgf}
\usepackage{lmodern}
\usepackage{import}
\usepackage{xr}
\usepackage{enumitem}
\usepackage{soul}
\usepackage[normalem]{ulem}
\usepackage{multirow}
\usepackage{tabularx,ragged2e,booktabs,caption}
\UseRawInputEncoding
\usepackage{xcolor}
\usepackage{graphicx}
\usepackage{lipsum}

\newif\ifhyper
\hypertrue
\ifhyper
\hypersetup{
   citecolor = {green},
   colorlinks = {true}, 
   urlcolor = {blue} 
}
\fi

\newcommand{\beq}{\begin{equation}}
\newcommand{\eeq}{\end{equation}}
\newcommand{\beqa}{\begin{eqnarray}}
\newcommand{\eeqa}{\end{eqnarray}}

\newcommand{\comment}[1]{}

\def\Longarrow{\protect\@lra}
\def\@lra{\relbar\joinrel\relbar\joinrel\relbar\joinrel%
          \relbar\joinrel\rightarrow}

\def\norm#1{\left\lVert#1\right\rVert}

\pgfplotsset{compat=1.18}

\begin{document} 

\title{Blockchain Network Analysis \\ using Quantum Inspired Graph Neural Networks \& Ensemble Models}

\author{Luigi D'Amico}
\author{Daniel De Rosso}
\author{Ninad Dixit}
\author{Raul Salles de Padua}
\author{Samuel Palmer}
\author{Samuel Mugel}
\author{Rom\'{a}n Or\'{u}s}

\affiliation{Multiverse Computing, Parque Cientifico y Tecnol\'{o}gico de Gipuzkua, Paseo de Miram\'{o}n, 170 3$^{\,\circ}$ Planta, 20014 Donostia / San Sebasti\'{a}n, Spain}

\author{\\ Holger Eble}
\author{Ali Abedi}
\affiliation{Innovation Hub - Bundesdruckerei GmbH, Kommandantenstraße 18, 10969 Berlin, Germany}




\begin{abstract}
In the rapidly evolving domain of financial technology, the detection of illicit transactions within blockchain networks remains a critical challenge, necessitating robust and innovative solutions\cite{Weber2019}. This work proposes a novel approach by combining Quantum Inspired Graph Neural Networks (QI-GNN) with flexibility of choice of an Ensemble Model using QBoost\cite{Qboost2012} or a classic model such as Random Forrest Classifier\cite{Gilles2015}. This system is tailored specifically for blockchain network analysis in anti-money laundering (AML) efforts. Our methodology to design this system incorporates a novel component, a Canonical Polyadic (CP) decomposition layer\cite{Hua2022} within the graph neural network framework, enhancing its capability to process and analyze complex data structures efficiently.

Our technical approach has undergone rigorous evaluation against classical machine learning implementations, achieving an F2 score of 74.8\% in detecting fraudulent transactions. These results highlight the potential of quantum-inspired techniques, supplemented by the structural advancements of the CP layer, to not only match but potentially exceed traditional methods in complex network analysis for financial security. The findings advocate for a broader adoption and further exploration of quantum-inspired algorithms within the financial sector to effectively combat fraud.

\end{abstract}

\maketitle

\section{Introduction}
\label{sec:intro}

Money laundering poses a significant global challenge, with an estimated \$715 billion to \$1.87 trillion, equivalent to 2-5\% of the global GDP, being laundered annually\cite{Alarab2020}\cite{UNreport}. This illicit activity involves concealing the origins of illegal funds through various means such as casino transactions, real estate purchases, and overvaluing legitimate invoices. Typically, money laundering comprises three main stages: placement, layering, and integration. Placement involves introducing dirty money into the financial system, layering entails complex transactions to obfuscate the funds' source, and integration involves withdrawing laundered proceeds for legitimate use. 

To combat money laundering, anti-money laundering (AML) measures are implemented, primarily focusing on regulatory compliance within financial institutions. These measures include Know Your Customer (KYC) standards, transaction monitoring, account restrictions, and the submission of Suspicious Activity Reports (SARs) to law enforcement agencies. A five-step process is commonly employed, involving compliance training, KYC procedures, transaction monitoring systems (TMS), manual review of flagged activities, and filing SARs. 

Transaction monitoring systems utilize rules-based threshold protocols to analyze transaction data and identify suspicious activities. Two main topologies illustrate how money laundering occurs: one involving intermediary accounts forming a column, and the other featuring a row of intermediary accounts passing funds sequentially to the receiver account\cite{Karim2023}. Constructing money transfer graphs aids in visualizing transactions, where nodes represent accounts and edges represent transactions. However, identifying money laundering from these graphs is challenging due to the complexity of transactions and the ability of criminals to mask their activities. 

Recent approaches \cite{Weber2019} have integrated non-graph-based models with graph neural networks (GNNs) to enhance detection capabilities. Combining graph analytics with tree-based ensemble models enables the modeling of spatial and temporal information in large-scale transaction graphs. By treating money laundering detection as a node classification problem, nodes exhibiting anomalous characteristics can be identified as potential money launderers. Semi-supervised graph learning techniques are applied to financial transaction graphs to accurately identify nodes involved in potential money laundering activities.

This paper is organized as follows: In Section.~\ref{sec:problem}, we provide an overview of the illicit transaction detection problem that we tackle in the Elliptic dataset. In Section.~\ref{sec:methods}, we introduce the main concepts of tensor networks and present our methodology for building a tensorized GNN (tGNN) + Ensemble classifier model. In Section.~\ref{sec:experimental_setup}, we discuss the architectures of the tGNN models, the training procedure and the quality metrics we used for assessing the accuracy and performance of the models. Results of our study are presented and discussed in Section.~\ref{sec:results}. Finally, Section.~\ref{sec:conclusion} is devoted to the discussion and conclusions.

\section{Problem overview}
\label{sec:problem}

\subsection{Anti-Money Laundering (AML) Background}
\label{sec:problem_background}

The growth of cryptocurrency adoption has brought with it challenges related to illicit financial activities. This makes anti-money laundering (AML) regulations critical for maintaining the integrity of financial systems, yet they pose significant costs for financial institutions and inadvertently contribute to financial exclusion for marginalized groups\cite{Alarab2020}. The advent of cryptocurrencies has introduced a paradox: while the pseudonymity of cryptocurrencies allows money laundering to operate under the radar, the transparency of blockchain technology provides investigators with unprecedented access to transaction data. This dual nature presents an opportunity to enhance AML efforts leveraging advanced machine learning techniques.

Cryptocurrencies have transformed the financial landscape by enabling peer-to-peer transactions without intermediaries. Despite the benefits, the anonymity associated with cryptocurrency transactions has made them attractive for illicit activities, such as money laundering, terrorism financing, and other forms of financial crime. Consequently, robust AML measures are essential to mitigate these risks. The open and transparent nature of blockchain technology, which records all transactions in a public ledger, offers a unique advantage for AML efforts by allowing detailed transaction tracking and analysis.

Recent advances in machine learning and artificial intelligence have shown promising enhancing AML capabilities\cite{Weber2019}. Considering those challenges, we propose the integration of graph neural networks (GNNs) and quantum-inspired classifiers into AML systems represents a cutting-edge approach to detecting and preventing illicit transactions within blockchain networks. GNNs are especially suited to the graph-like structure of blockchain data, capturing relational information among transactions to identify suspicious patterns. Meanwhile, quantum-inspired techniques like QBoost\cite{Qboost2012} leverage quantum computing principles to optimize the selection and performance of weak classifiers, enhancing the overall accuracy and robustness of AML models.

The Elliptic Data Set\cite{Eelliptic}, a comprehensive dataset of Bitcoin transactions, provides a valuable resource for developing and evaluating these advanced AML systems. Comprising over 200,000 transactions and enriched with 166 features per node, this dataset facilitates the application of both traditional machine learning models and novel graph-based approaches. The dataset's rich feature set, which includes both local and aggregated information, allows for a multifaceted analysis of transaction behavior, improving the detection of illicit activities. Details in Section~\ref{sec:experimental_setup} of this paper.

This paper explores the architecture and performance of a blockchain-based AML system utilizing GNNs and quantum-inspired classifiers. We assess the system using standard evaluation metrics such as Precision, Recall, F1, and F2 scores, offering a comprehensive view of its effectiveness. Our aim is to contribute to the ongoing efforts in AML research by demonstrating the potential of integrating advanced machine learning techniques with blockchain technology, ultimately enhancing financial security while promoting financial inclusion.

\subsection{Problem Formulation}
\label{sec:problem_formulation}

In this study, we consider a financial transaction network represented by a graph $G = (V,E)$, where $V$ denotes accounts and $E$ represents the monetary transfers between these accounts. The vertex set $V$ is partitioned into three subsets: $X$, $W$, and $Y$. Here, $W$ comprises internal bank accounts, while $X$ and $Y$ represent external accounts with net money transfers into and out of the bank, respectively. An edge $(i,j) \in E$ implies a transfer of money from account $v_{i}$ to $v_{j}$ for $v_{i} ,v_{j} \in V$ and $e_{ij}$ as the transfer amount. 

To effectively analyze this graph, we employ knowledge graph techniques, modeling the network as triplet facts $(h,r,t) \in F$, thereby expanding our graph representation to $G=(V,E,F)$. This framework identifies relationships $r \in R$ between head $h \in V$ and tail $t \in E$ within the transactions. 

We apply a semi-supervised learning approach to detect suspicious activities within the network\cite{Karim2023}. Initially, we embed each account $v_{i}$  into a lower-dimensional vector space $R_{d}$, using an embedding model $\Gamma$, resulting in a set of embedding vectors $\vec{v_{i}}$. Each account $v_{i}$ is represented by a vector $\vec{v_{i}}$, where $d$ is the embedding dimension, and $N$ is the total number of accounts.

The core of our method involves training a binary classifier $f$ on $\vec{V}$ to predict potential suspicious activities. For each node $i$, the classifier outputs $\hat{y}_{i} = f(v_{i})$ = 1 if the account is flagged (e.g., for suspicious activity reports or illicit transactions), and 0 otherwise. 

To reinforce the learning process under a semi-supervised paradigm, we run experiments excluding a percentage of nodes and their connections, training the graph embedding (GE) model on this reduced sub-graph. During inference, we reintroduce these nodes, generate their embeddings using the previously trained GE model, and predict their labels. This semi-supervised technique helps us to handle unlabeled data effectively while improving the model's ability to identify suspicious nodes in financial networks.

\section{Related work} 
\label{sec:methods}

In this section, we describe how to use quantum-inspired tensor network methods to improve the efficiency of GNN architectures\cite{Wang2023}, \cite{Baghershahi2023}, \cite{Jia2020}, and \cite{Zhao2023}. After introducing basic concepts of tensor networks, we present a methodology for constructing a tensorized Graph Neural Network (tGNN) and describe how the number of parameters is reduced compared to a GNN.

\subsection{Graph Neural Networks (GNNs)}
The exploration of deep learning methods for graph-structured data has seen an exponential increase in interest due to the unique combinatorial complexity these structures present. Among the varied strategies developed to handle large-scale graphs, Graph Convolutional Networks (GCNs) have shown significant promise in making efficient use of graph topology and node features\cite{Rozemberczki2021}\cite{knuthwebsite}.  

GCNs function by utilizing layers of graph convolution that extend the principles of traditional perceptrons through a neighborhood aggregation process, inspired by spectral convolution. A typical setup includes an adjacency matrix $A$ and node embedding matrix $H^{(l)}$, which undergo transformations through a weight matrix $W^{(l)}$ to produce updated node embeddings $H^{(l+1)}$. This process is mathematically represented as $$H^{(l+1)} = \sigma(\tilde{A}H^{(l)}W^{(l)})$$ where $\tilde{A}$ is a normalized version of $A$, refined for spectral graph filtering. 

In a standard implementation, a two-layer GCN can be described by the equation: $$H^{(2)} = \text{softmax}(\tilde{A} \cdot \text{ReLU}(\tilde{A}XW^{(0)})W^{(1)})$$. This model also incorporates a skip connection variant, known as Skip-GCN, which enhances feature propagation by adding a connection from initial node features directly to the output, formulated as $$H^{(2)} = \text{softmax}(\tilde{A} \cdot \text{ReLU}(\tilde{A}XW^{(0)})W^{(1)} + XW^{(1)})$$. This configuration allows the Skip-GCN to potentially exceed or at least match the predictive capability of logistic regression, as when $W^{(0)}$ and $W^{(1)}$ are null, the model simplifies to logistic regression.

Expanding the application of GCNs to temporal data, particularly in financial systems where transactions are inherently time-stamped, introduces the concept of EvolveGCN\cite{Pareja2019}. This model adapts the GCN framework to capture dynamics over time by employing a separate GCN for each time-step, interconnected through a recurrent neural network (RNN)\cite{Khrulkov2019}. This architecture allows the model to evolve, leveraging past system states to inform future predictions, thus enhancing its ability to generalize across different time periods. The dynamic nature of financial systems, with evolving undercurrents of transactions, can be more accurately modeled using this temporal approach. In practice, EvolveGCN uses the embeddings of the top influential nodes at each time-step as input, updating the system state continually with an RNN mechanism, such as a Gated Recurrent Unit (GRU).

\subsection{Canonical Polyadic (CP) Decomposition Layer GNN Implementation}

In the context of enhancing Graph Neural Networks (GNNs) for better representation and modeling of high-order interactions, the tensorized Graph Neural Network (tGNN) introduces the symmetric CP decomposition to significantly extend the capability of GNNs \cite{Hua2022}. This advancement is particularly aimed at capturing the non-linear and multiplicative interactions among node features, which are crucial for complex graph-based tasks but are not efficiently handled by traditional aggregation functions such as sum, mean, or max.

The CP layer employed in tGNN efficiently parameterize permutation-invariant multi-linear maps, enabling the network to process and integrate high-dimensional feature interactions that are inherently nonlinear. This approach not only facilitates a deeper and more nuanced learning of graph structures but also addresses the scalability issues typically associated with modeling high-order interactions, thanks to the decomposition nature of the CP layer. The layer's ability to compute any permutation-invariant multi-linear polynomial is a notable theoretical advantage, providing it with a foundation to surpass the expressive power of conventional pooling methods used in GNNs. 

\begin{figure}[t!]
\centerline{\includegraphics[width=8cm]{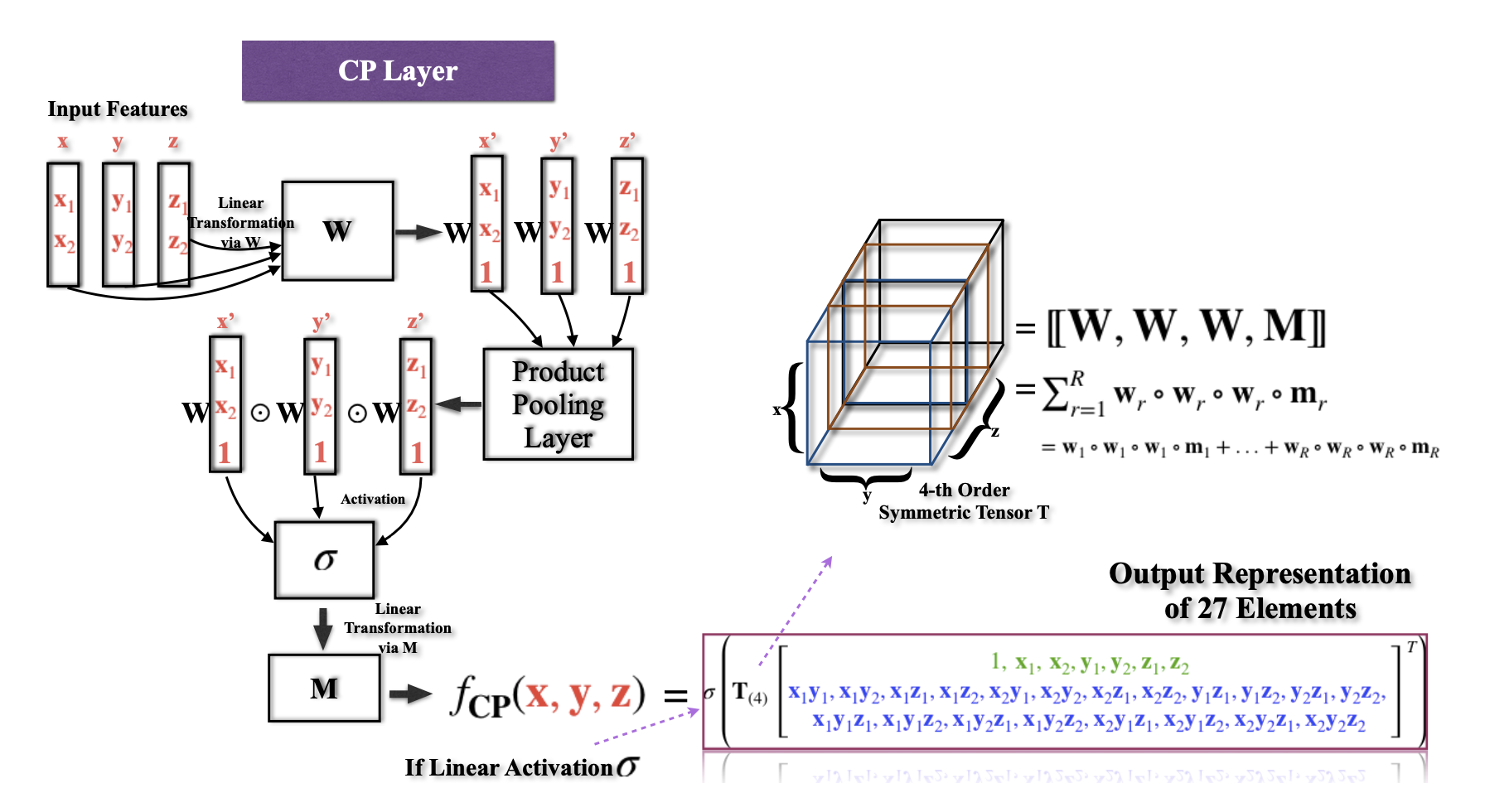}}
\caption{Extracted from \cite{Hua2022}: on the left side is the Sum pooling followed by a FC layer. On the right, the CP layer can be interpreted as a combination of product pooling with linear layers (with weight matrices W and M) and non-linearities.} 
\label{Fig:cp_layer}
\end{figure}

Thus, the tensorized Graph Neural Network (tGNN) leverages the symmetric CP decomposition to enable highly expressive aggregation operations in GNNs. The key advancement here is the efficient parameterization of permutation-invariant multi-linear maps that capture high-order interactions among node features. The principal formula presented \[f(x_1, ..., x_k) = T \times_1 x_1 \times_2 ... \times_k x_k\] which describes how the CP layer aggregates node features by contracting a tensor $T$ along multiple modes with feature vectors $x_1, ..., x_k$. This tensor $T$ is parameterized by a partially symmetric CP decomposition, which preserves permutation invariance and ensures scalability. The effectiveness of tGNN is attributed to its ability to handle complex, high-dimensional data interactions beyond the capabilities of typical sum or mean pooling layers, making it suitable for intricate graph-based learning tasks. The section underscores tGNN's potential by detailing its theoretical underpinnings and practical implications in enhancing the expressiveness and performance of graph neural networks. FIG.~\ref{Fig:cp_layer}. illustrates the CP pooling layer and compares it with sum pooling.

Empirically, tGNN has demonstrated superior performance on several node and graph classification benchmarks against existing state-of-the-art methods. This indicates its practical effectiveness in leveraging the enriched feature interactions for more accurate predictions. The use of CP decomposition thus represents a significant step forward in the design of graph neural architectures, promising enhancements in both the expressiveness and functional depth of GNNs for various applications.

\subsection{Ensemble Models}
In the ensemble models domain, particularly focusing on ensemble tree boosting systems, recent advancements have significantly enhanced the scalability and effectiveness of these methods for machine learning challenges. Chosen literature \cite{Gilles2015}\cite{Chen2016} discuss two such systems, XGBoost and a general exploration of ensemble methods.

XGBoost stands out due to its scalability and superior performance on many machine learning tasks. It employs a gradient boosting framework that optimizes both computational efficiency and memory usage, enabling it to handle large datasets effectively. Key innovations within XGBoost include a novel sparsity-aware algorithm that optimizes handling of sparse data, and a weighted quantile sketch, which is essential for managing instance weights in large datasets. The system is designed to scale across multiple cores and nodes efficiently, thus facilitating rapid model exploration and iteration.

\begin{figure}[t!]
\centerline{\includegraphics[width=8cm]{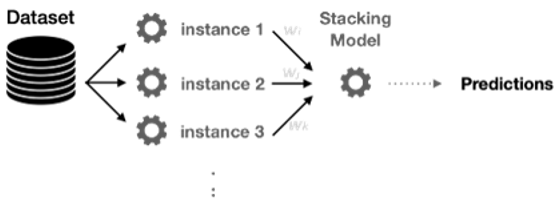}}
\caption{QBoost: Quantum version of an ensemble algorithm. QBoost builds an initial ensemble model of “weak” classifiers to then select optimal subset of classifiers by solving a hard optimization problem} 
\label{Fig:ensemble_models}
\end{figure}

Ensemble models, on a broader concept, emphasize the integration of multiple models to improve predictive performance. These ensemble methods often leverage multiple learning algorithms to obtain better predictive performance than could be obtained from any of the constituent learning algorithms alone. This approach is beneficial in various applications, including but not limited to, anomaly detection, risk management, and other areas requiring robust predictive capabilities. See FIG~\ref{Fig:ensemble_models}.

Ensemble models provide insights into their practical applications and theoretical underpinnings, illustrating their pivotal role in advancing machine learning and predictive analytics.

\subsection{QBoost}
The concept of QBoost emerges from the integration of quantum computing techniques into classical machine learning models, specifically ensemble learning methods\cite{Qboostgithub24}\cite{Qboost2012}. QBoost is fundamentally designed to optimize decision tree ensemble methods by employing a quantum annealer to select the best weak learners from an extensive set of candidate weak learners.

At the mathematical core of QBoost is the formulation of the problem as a quadratic unconstrained binary optimization (QUBO), which is particularly suitable for processing on quantum annealers. The main objective in QBoost is to minimize both the misclassification error and the model complexity by finding an optimal subset of weak classifiers. This is achieved by expressing the model selection problem as minimizing a weighted sum of the misclassification errors of the individual classifiers, along with a regularization term that penalizes the ensemble's complexity to avoid overfitting. It is mathematically represented as \[H_Q(w)=\sum_{s}\left(\frac{1}{N}\sum_{i}^{N}w_ih_i(\overrightarrow{x_s})-y_s\right)^2 + \lambda\norm{w}_0\]

\begin{figure}[t!]
\centerline{\includegraphics[width=8cm]{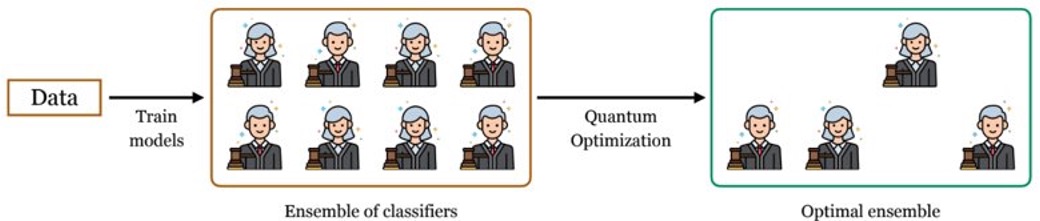}}
\caption{ QBoost: Quantum version of an ensemble algorithm. QBoost builds an initial ensemble model of weak classifiers to then select optimal subset of classifiers by solving a hard optimization problem} 
\label{Fig:qboost}
\end{figure}

The weak learners in QBoost are typically decision trees, chosen due to their effectiveness in handling diverse datasets and their capability to model complex decision boundaries. Each tree in the ensemble votes on the classification outcomes, and the ensemble's final output is determined by a weighted majority vote of these weak learners. The key innovation in QBoost is in how these weights and the selection of trees are determined. Instead of relying on classical iterative methods like gradient boosting, QBoost leverages quantum annealing to efficiently explore the solution space and identify the most effective combination of trees, thus potentially achieving better performance especially in scenarios where classical methods might struggle with the combinatorial nature of the problem. See FIG ~\ref{Fig:qboost}.

This approach not only harnesses the computational power of quantum devices but also aligns with advancements in ensemble methods by integrating quantum optimization techniques to refine the selection process of weak learners, potentially leading to more accurate and robust predictive models.
\section{Methodology} 
\label{sec:experimental_setup}

In this section, we describe the dataset, experimental setup we used to build, train and test the different alternatives to assemble a system with different model components for detecting illicit transactions.

\subsection{Data}
The Elliptic Dataset\cite{Eelliptic} serves as a foundational component in constructing our system designed to identify illicit transactions within the Bitcoin network. This dataset is distinguished by its unique composition, including over 200,000 Bitcoin transactions represented as nodes and 234,355 directed payment flows depicted as edges between these nodes. These transactions span 49 distinct time steps, with temporal details estimated based on when transactions were confirmed by the network.

Each transaction node in the dataset is enriched with 166 features, which fall into two categories: local and aggregated features. The first 94 local features provide immediate information about the transaction, such as the time step, the number of inputs and outputs, transaction fees, and output volume. These features also encapsulate aggregated statistics like the average Bitcoin received or spent by the inputs and outputs, and the average number of incoming or outgoing transactions associated with these entities.

\begin{figure}[t!]
\centerline{\includegraphics[width=8cm]{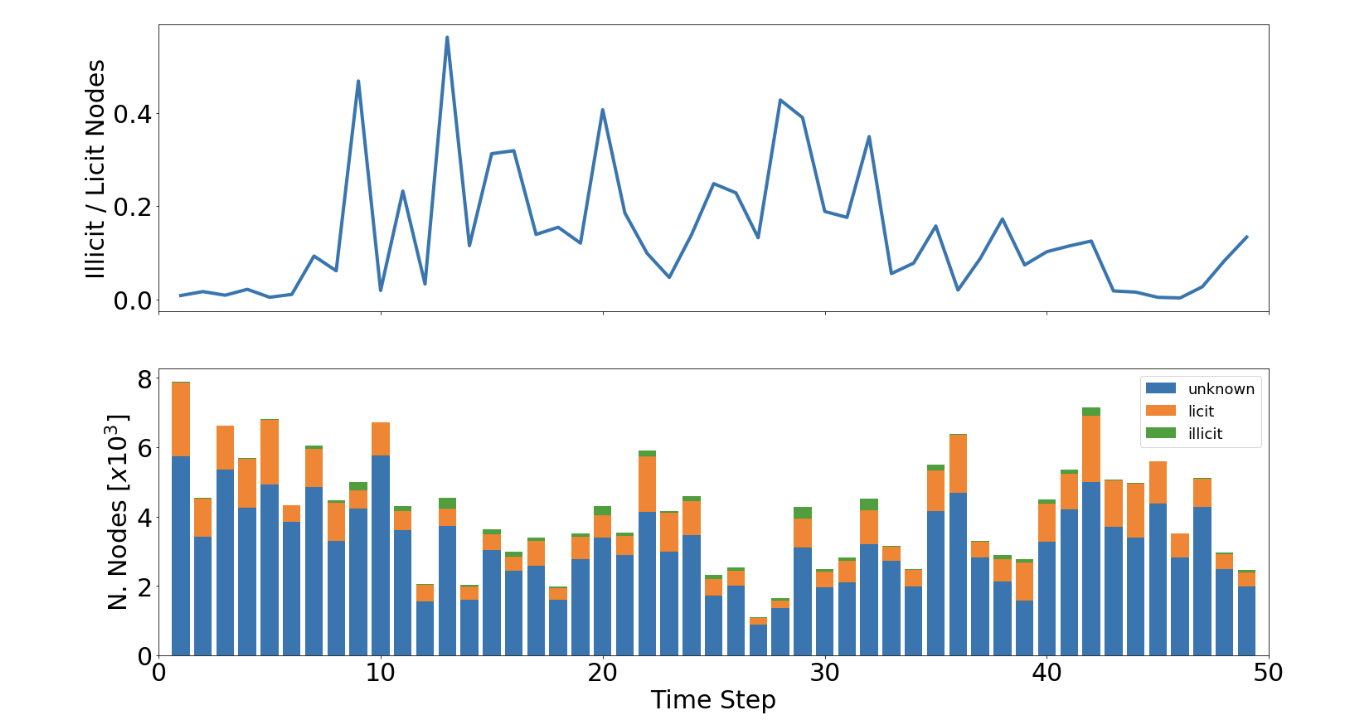}}
\caption{Extracted from \cite{Weber2019}: (Top) Fraction of illicit vs. licit nodes at different time steps in the data set. (Bottom) Number of nodes vs. time step.} 
\label{Fig:elliptic}
\end{figure}

The additional 72 features are aggregated, crafted by gathering data one-hop backwards and forwards from the central node in question. These features include statistical measures such as the maximum, minimum, standard deviation, and correlation coefficients of the neighboring transactions. This dual-feature framework allows for a robust analysis of not only the transactions themselves but also their broader context within the Bitcoin network, capturing both immediate and extended network effects.
The labeling of nodes within this dataset reflects a binary classification of transactions into "licit" or "illicit," with labels derived from connections to identified legal or illegal entities, respectively. Not all transactions are labeled; many remain unclassified yet still carry the detailed features that contribute to the network's overall analysis. See FIG.~\ref{Fig:elliptic}. 

This dataset's comprehensive and detailed structure makes it an exemplary resource for training machine learning models to recognize patterns indicative of illicit activities. It provides a rich ground for experimenting with various models, including Graph Convolutional Networks, which are particularly suited to leveraging the relational data inherent in graph-structured data like the Bitcoin transaction network.

\begin{figure}[t!]
\centerline{\includegraphics[width=8cm]{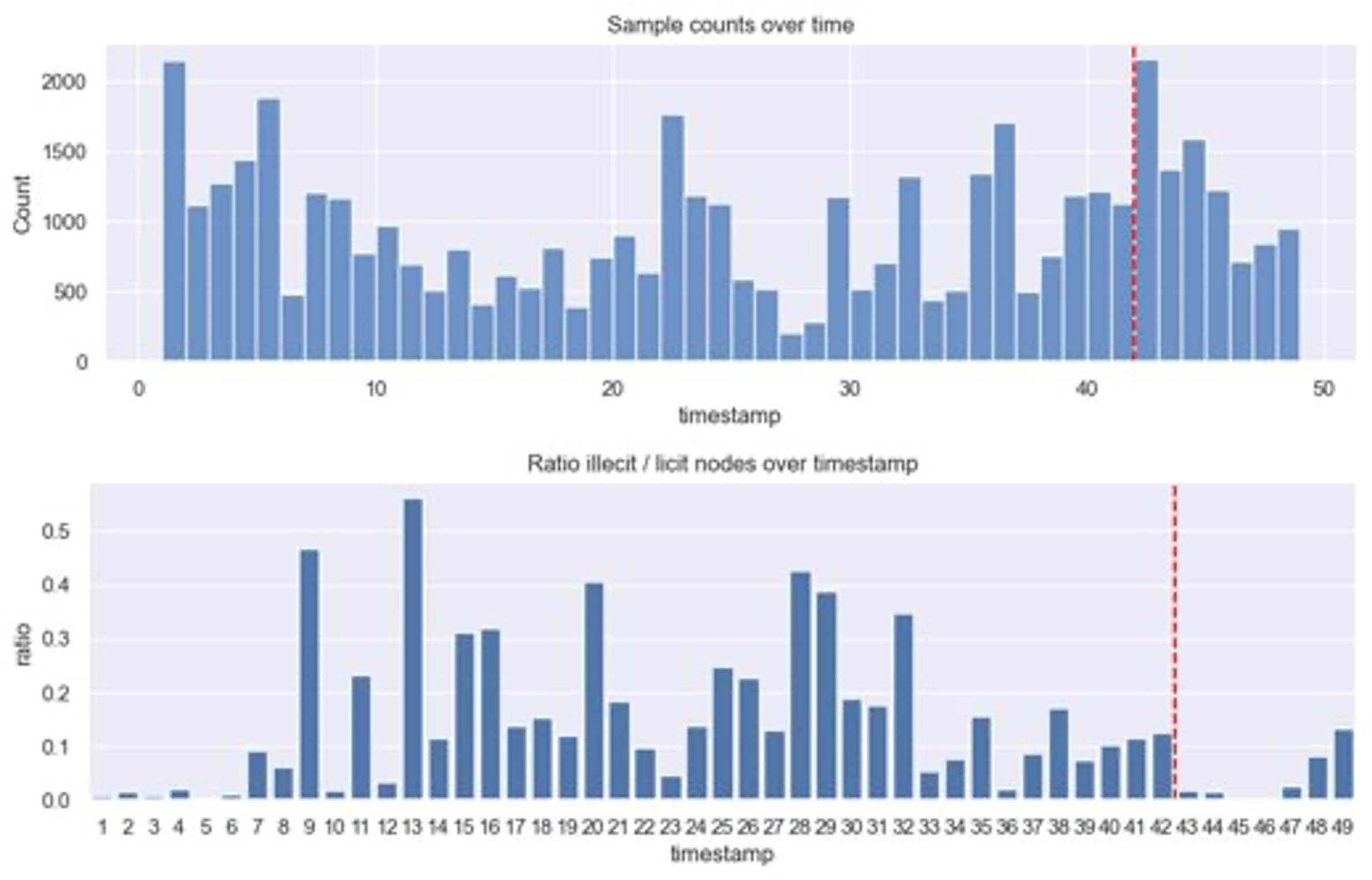}}
\caption{The Elliptic dataset is graph representing Bitcoin transaction that are classified licit or illicit. (Top) A time stamp is associated with each node (49 in total). (Bottom) Emerging Dark Market Shutdown event occurring at time step 43 identified.} 
\label{Fig:dark_market}
\end{figure}

An exploratory data analysis was conducted. A time stamp is associated with each node (49 in total). There are no edges connecting different time steps. We have different graph for each time steps. In the dataset there is a Dark Market Shutdown emerging event occurring at time step 43. See FIG.~\ref{Fig:dark_market}. This event causes the system to behave differently as a consequence.

We use a train, validation, test split. We split the sets in considering the temporal information because it reflect the nature of the data. In particular we adopt the first 29 time steps as train sets, the subsequent 5 time steps as validation set, and the remaining 15 time steps as test set. The train set is directly used for the models training. The validation is used for hyperparameters tuning. Once we pick the best performing model on validation set, we evaluate the performance of the model on the test set. What we will report are the metrics on the test set.

\subsection{System architecture}

\begin{figure}[t!]
\centerline{\includegraphics[width=8cm]{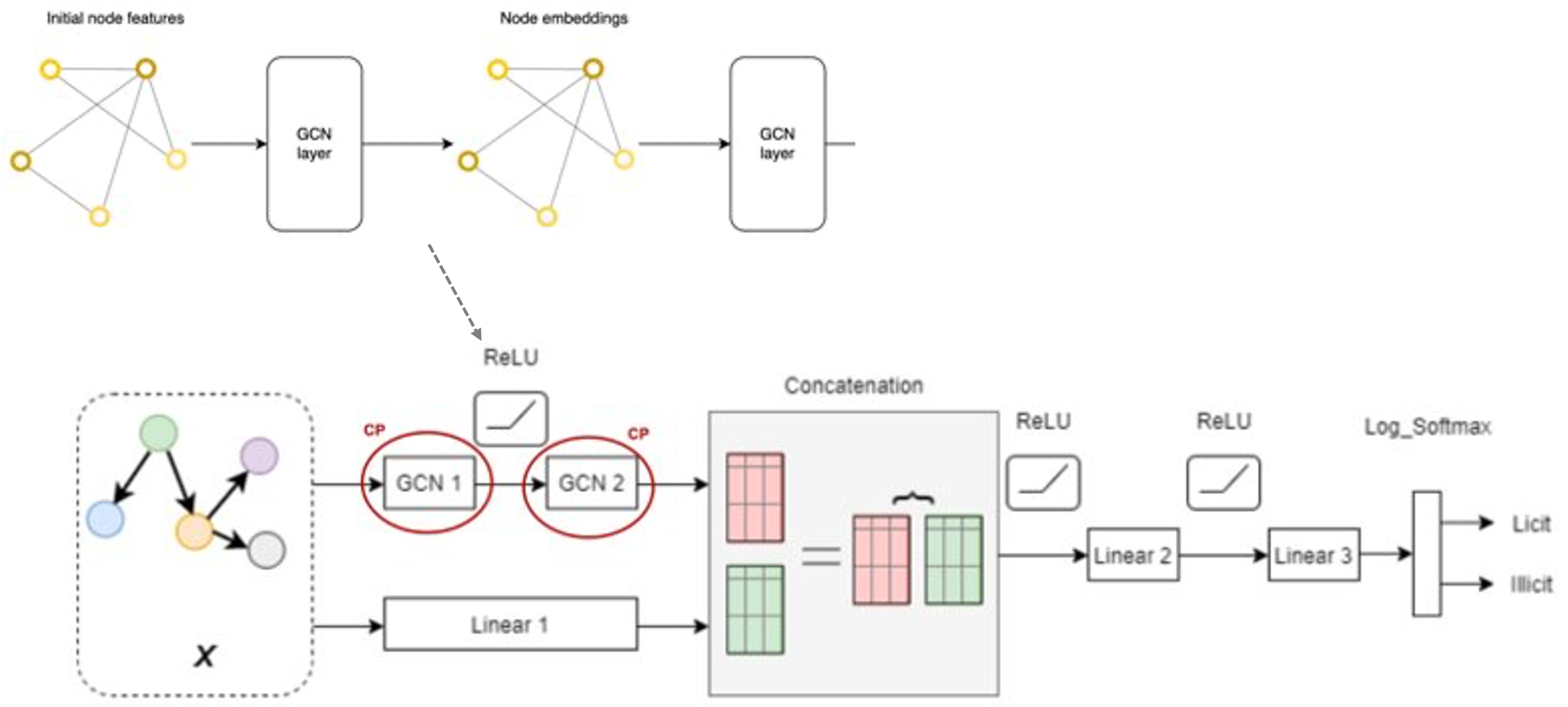}}
\caption{GCN intertwined with linear layers. Can be viewed as two sub-models: (Top) Graph-based approach. (Bottom) Based on linearly transformed feature matrix, to reuse of the latent features and maximize the flow of information} 
\label{Fig:GCN}
\end{figure}

\begin{figure}[t!]
\centerline{\includegraphics[width=8cm]{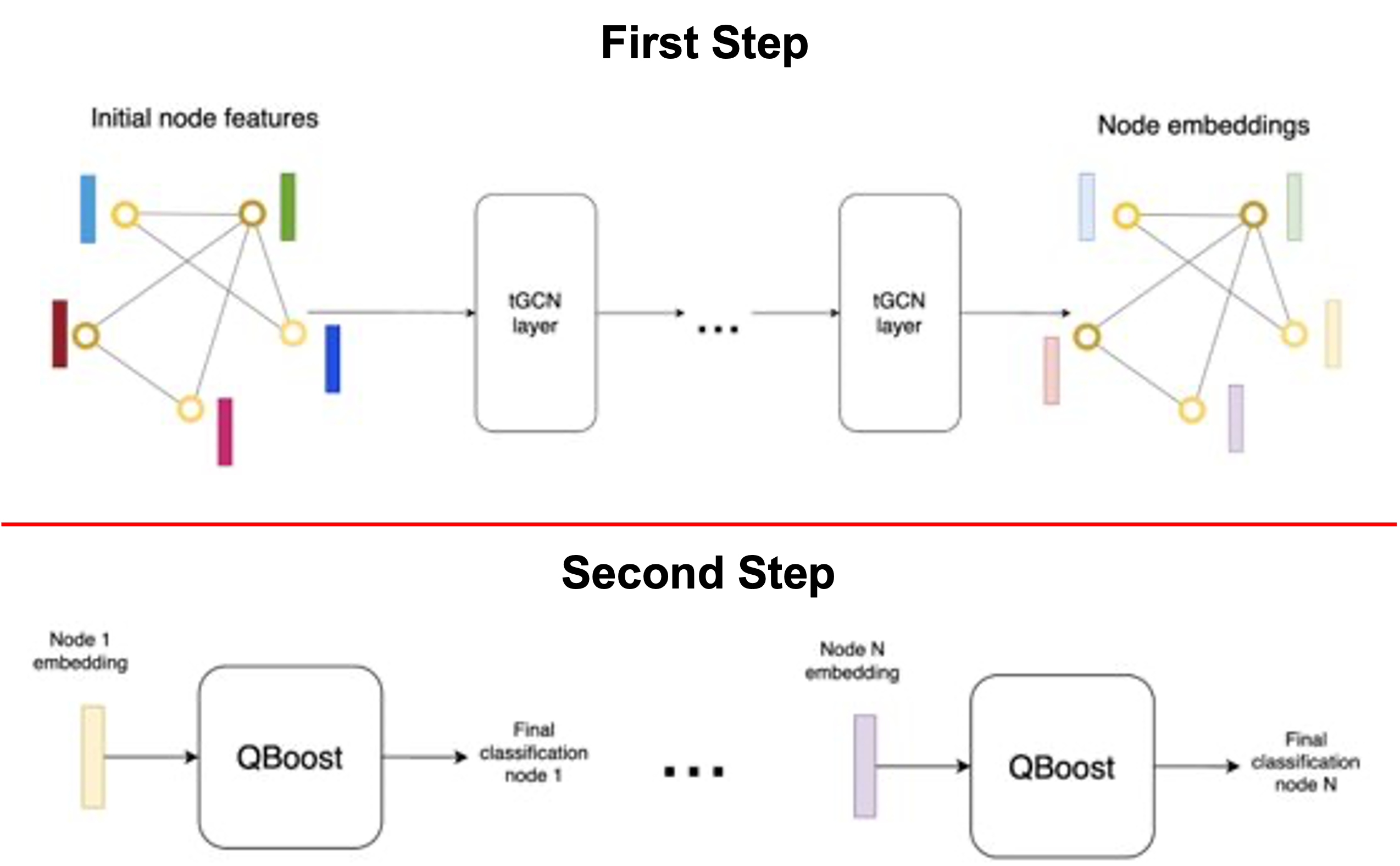}}
\caption{First step: leverage tGCN to generate the embedding representation of each node. Second step: QBoost to generate the final classification for each node. 
QBoost: Quantum ML algorithm that utilizes quantum principles to enhance classical boosting algorithm} 
\label{Fig:QI_full}
\end{figure}

During the design of our system architecture for a blockchain-based anti-money laundering (AML) solution, we considered three primary approaches, each taking advantage of distinct machine learning paradigms optimized for the unique structure and features of our dataset:

\begin{enumerate}
    \item Graph Neural Networks (GNNs): The inherent graph structure of the blockchain dataset, which includes nodes representing transactions and edges depicting payment flows, lends itself well to the application of GNNs to embed information into the system. These networks are highly effective in processing the relational information embedded within the graph, as they not only analyze individual node features but also incorporate contextual information from adjacent nodes. This capability is crucial for accurately classifying and detecting patterns indicative of illicit transactions within the network.
    \item Classical Machine Learning Models: Traditional machine learning models remain relevant due to their proven effectiveness across diverse datasets, including those with complex aggregated features. In this approach, we consider leveraging robust tree-based ensemble methods such as Random Forest and XGBoost. These models are particularly adept at handling structured data and are valued for their high accuracy, interpretability, and ability to handle both linear and non-linear relationships without the need for extensive pre-processing.
    \item Hybrid GNN and Machine Learning Models: By harnessing the strengths of both GNNs and classical machine learning, this approach involves training a GNN to capture and encode the complex structural relationships within the graph into lower-dimensional embeddings. These embeddings, which effectively summarize the transactional context and network dynamics, can then be used as enriched feature inputs for traditional machine learning models. This hybrid model aims to combine the relational learning capabilities of GNNs with the predictive power of established machine learning techniques to enhance overall model performance.
\end{enumerate}
Additionally, to further increase the predictive accuracy and robustness of these models, we explore the integration of quantum-inspired approaches such as QBoost and Tensor Neural Networks (TNNs). These quantum methods are designed to optimize the ensemble model selection process and to enhance feature interaction modeling, respectively. See FIG.~\ref{Fig:GCN} for a CP layer inclusion in the GNN and FIG.~\ref{Fig:QI_full} for the full Quantum Inspired System. The inclusion of these advanced techniques not only can improve the detection capabilities of our system as shown, but also provide flexibility of proper choosing the best alternative depending on technical requirements of our system, providing a cutting-edge solution to the challenge of identifying and preventing illicit activities in cryptocurrency transactions.

\subsection{Graph Neural Network (GNN) and tensorized Graph Neural Network (tGNN) Training setup}
We implemented a GNN architecture and its tensorized version, as shown in FIG.~\ref{Fig:cp_layer} and FIG.~\ref{Fig:GCN} respectively, to compare benchmark the performance of these two architectures. For each option, we trained multiple models using different hyperparameters. For each hyperparameter configuration we trained 5 models with different random seeds in order to mitigate randomness factors. The metric we used for comparing different experiments was the F2 score. The best hyperparameter configuration was selected aggregating performance (F2 score) for different random seeds on each hyperparameter configuration and checking the mean value on the validation set. We trained the models for 500 epochs since we have seen that is enough to bring the model to convergence \cite{Jia2020},\cite{Pareja2019}, and \cite{Zhao2023}. We used an early-stop approach, meaning that we consider the model that has the best performance in terms of the loss value on the validation set during the training to prevent overfitting. For training GNNs we only used the local features, 93 in total, because the Graph Convolutional Layers are able to process the information of adjacent node features leveraging the graph topology themselves \cite{Weber2019}. The training procedure is the following. Since we have independents graphs divided by time-steps, for each epoch, we do the forward pass for each time-steps independently and sum up the losses. When the training graphs are finished, we back-propagate and update all weights. The chosen loss function is Cross Entropy, commonly used in deep learning classification tasks using a label weights of $0.7$ and $0.3$ respectively for the illicit and licit labels. This approach is semi-supervised. The entire dataset (labeled and unlabeled nodes) is used to feed-forward, since we can leverage the features of unlabeled nodes, but when it is time to compute the loss, we only use the prediction and the true label of labeled nodes. Regarding the chosen optimizer, we used Adam otimizer with a learning rate of $0.001$ and a weight decay of $5 \text{x} 10^{-5}$. The classification threshold is chosen in an automatized fashion: it is the threshold that maximize the f2 score on the validation set.

\begin{table*}[ht]
\centering
\begin{tabular}{|c|c|c|c|c|c|c|c|}
\hline
Model & F2 score & F1 score & Precision & Recall & n. parameters & Train time (s) & Inference time (s) \\
\hline
GCN based & $0.607 \pm 0.005$ & $ 0.530 \pm 0.028 $ & $0.439$ & $0.675$ & $11902$ & $77.61 \pm 6.99$ & $0.055 \pm 0.004$  \\
CP-GCN based & $0.610 \pm 0.041$ & $0.500 \pm 0.066$ & $0.387$ & $0.721$ & $4031$ & $68.61 \pm 2.75$ & $0.040 \pm 0.002$ \\
\hline
\end{tabular}
\caption{GNN against tGNN results (means and std. dev. on test set is reported)}
\label{table:resultsGCN}
\end{table*}

\begin{table*}[ht]
\centering
\begin{tabular}{|c|c|c|c|c|c|c|c|}
\hline
Model & F2 score & F1 score & Precision & Recall & n. estimators & Train time (s) & Inference time (s) \\
\hline
Random Forest & $0.679 \pm 0.016$ & $0.780 \pm 0.026$ & $0.878$ & $0.705$ & $50$ & $0.58 \pm 0.01$ & $0.33 \pm 0.010$ \\ 
XGBoost & $0.706$ & $0.689$ & $0.647$ & $0.737$ & $50$ & $0.07 \pm 0.36$ & $0.365 \pm 0.005$ \\ 
QBoost & $0.726 \pm 0.008$ & $0.703$ & $0.689$ & $0.717$ & $42.7 \pm 1.4$ & $47.32 \pm 2.02$ & $ 0.25 \pm 0.004$ \\ 
\hline
\end{tabular}
\caption{Ensemble algorithms results (means and std. dev. on test set is reported)}
\label{table:resultsEnsemblers}
\end{table*}

\subsection{Ensemble Classifier Training setup}

We consider different model classes, Random Forest\cite{Gilles2015}, XGBoost\cite{Chen2016} and a QBoost version developed for this project. In the GNN, we trained multiple models for each model class using different hyperparameters configuration and different random seeds. We then picked the model that possessed the best F2 score on the validation set. For training these models, we use both the local features and the aggregated ones, resulting in total number of 166 features. The training approach is fully supervised, since we cannot leverage the information provided from the unlabeled nodes. The threshold selection is automatically settled for taking the best F2 score on the validation set, as we did in the case of GNN. 

\subsection{Performance metrics}

As we have an imbalanced dataset, performance metrics like accuracy can be misleading. For this reason, we measure model performance using precision, recall, F1, F2 and scores as \emph{evaluation metrics}. Precision, Recall, F1, and F2 scores are commonly used metrics for evaluating the performance of classification models, especially in scenarios where classes are imbalanced. Each of these metrics offers insights into different aspects of the system's performance, focusing on the balance between correctly predicted positive observations and the type of errors made (either false positives or false negatives). Here's an overview of each:
    \[Precision = \frac{\text{TP}}{\text{TP + FP}}\]
    where $TP$ is the number of true positives and $FP$ is the number of false positives.
    \[Recall = \frac{\text{TP}}{\text{TP} + \text{FN}}\]
    where $FN$ is the number of false negatives
    \[F1 = \frac{2 \cdot (\text{Precision} \times \text{Recall})}{\text{Precision} + \text{Recall}}\]
    The F1 score is the weighted average of Precision and Recall. Therefore, this score takes both false positives and false negatives into account. The F1 score is the harmonic mean of Precision and Recall.
    \[F_{\beta} = (1+\beta^{2})\frac{\text{Precision}\times\text{Recall}}{\beta^{2}\times\text{Precision}+\text{Recall}}\]
    \newline
    The $F_{\beta}$ score weights recall higher than precision customized to the value of $\beta$. It is used when the identification of false negatives is more important than the identification of false positives. Is this work we decided to use $\beta=2$, when the recall is weighted twice as heavily as precision, emphasizing the importance of minimizing false negatives.

\section{Results} 
\label{sec:results}

This section presents a detailed analysis of the results obtained from the implementation of a blockchain based anti-money laundering (AML) system. We leverage a combination of Graph Neural Networks (GNNs), its tensorized version specifically integrating a CP Decomposition Layer (tGNN) and quantum-inspired classifiers. The application of these advanced technologies aims to enhance the detection capabilities of illicit activities within cryptocurrency transactions. The evaluation metrics used to assess the system's performance include Precision, Recall, F1, and F2 scores, providing a multifaceted view of the model's effectiveness in identifying fraudulent transactions. 

During model training we chose F2 score as main metric to compare experiments, since we want to prioritize recall over precision. This analysis not only highlights the robustness and accuracy of the proposed model but also sheds light on the potential and limitations of integrating quantum-inspired methods in blockchain technologies for financial security.

\begin{table*}[ht]
\centering
\begin{tabular}{|c|c|c|c|c|c|c|}
\hline
Class & Model & F2 score & F1 score & Precision & Recall & Inference time (s) \\
\hline
Full classical & GCN + RF & $0.747 \pm 0.004$ & $0.789 \pm 0.021$ & $0.875 \pm 0.0589$ & $0.721 \pm 0.006$ & $0.39$ \\ 
Full quantum & CP-GCN + QBoost & $0.680$ & $0.618$ & $0.538$ & $0.728$ & $0.29$ \\ 
Hybrid & CP-GCN + RF & $0.748 \pm 0.004$ & $0.793 \pm 0.018$ & $0.882 \pm 0.052$ & $0.721 \pm 0.006$ & $0.37$ \\ 
\hline
\end{tabular}
\caption{Combined methods results (means and std. dev. on test set is reported)}
\label{table:resultsCombined}
\end{table*}

\begin{table*}[ht]
\centering
\begin{tabular}{|c|c|c|c|c|}
\hline
Model & F2 score & Train time (s) & Inference time (s) & Comment\\
\hline
CP-GCN & $0.610$ & $69$ & $0.04$ & Fastest prediction time \\ 
QBoost & $0.726$ & $47$ & $0.25$ & Compromise for F2 score and prediction time\\ 
XGBoost & $0.706$ & $0.07$ & $0.36$ & Fastest training time\\ 
CP-GCN + RF & $0.748$ & $70$ & $0.37$ & Highest F2 score\\ 

\hline
\end{tabular}
\caption{Selected most relevant architectures}
\label{table:resultsSelected}
\end{table*}

\subsection{System Results}

First, we present the results for the proposed Graph Neural Network (GNN) architecture versus its tensorized version (tGNN). For the GNN, as result of the hyperparameters optimization, layers' sizes were defined with $50$ and $30$ units respectively for the first and second linear layer and $50$ and $10$ for the first and second convolutional layers. Comparatively, for the tGNN, the linear layers sizes were defined with $25$ and $15$ units, and the convolutional layers sizes were $12$ and $5$. Regarding the ranks, they were set to $10$ and $4$.
As shown in Table \ref{table:resultsGCN}, the tensorized version achieves the same performance in terms of F2 scores with $66\%$ fewer parameters, improving training speed by $11\%$ and inference time by $27\%$. Despite the significant reduction in parameters, the same ratio is not observed for inference and training speed. This behavior is explained likely because of the computational intensive nature of the CP-layer used in the tGNN.

Next, the results of the Ensemble algorithm are shown in Table \ref{table:resultsEnsemblers} where, as result of the hyperparameter optimization, a total of $50$ weak-learners were defined for all the algorithms. We observe that QBoost achieves a better F2 score and faster prediction time compared to Random Forest and XGBoost. This suggests superior generalization capabilities of QBoost, attributable to its simplification of the initial set of ensemble models. The training time has drastically been increased, as expected, since QBoost training requires first training a set of weak learners and then solving an optimization problem to select the best subset.

Finally, the results of the system that combines such algorithms, which leverages embeddings generated by a neural network as input features for ensemble algorithms, are presented in Table \ref{table:resultsCombined}. The best F2 scores are obtained by the full-classical and hybrid pipelines. However, there is still a decrease in prediction time for a full-quantum pipeline.

Table \ref{table:resultsSelected} synthesizes the results of the most notable systems, each excelling in certain performance metrics. Depending on the most critical metric for the specific use case, one can select the most appropriate alternative.

\subsection{Analysis}
\label{sec:error_analysis}

Analyzing the results, we observe that both the classical and tensorized GNN architectures fail to achieve high performance in terms of F2 scores compared to the ensemble methods. This disparity is likely due to the heavy nonlinearity of patterns in the data, which makes ensemble methods more effective in such scenarios. It is important to note that utilizing ensemble methods to leverage the network topology requires manually crafting features that account for the graph structure. This process demands significant human effort and introduces the potential for errors.

On the other hand, GNN-based architectures offer the advantage of very fast inference, as they can predict the labels of all nodes simultaneously in a single forward pass. In contrast, ensemble methods must predict one node at a time, which is less efficient.

Additionally, combining the embeddings generated by the GNN with the modeling capacity of the ensemble method can lead to the development of more sophisticated pipelines. These combined approaches achieve better performance in terms of F2 scores but are inherently more computationally expensive.

\section{Conclusion and outlook}
\label{sec:conclusion}
We conducted an in-depth analysis utilizing state-of-the-art architectures for Graph Neural Networks (GNNs) and an ensemble algorithm specifically designed for anti-money laundering (AML) tasks. To benchmark the performance of these architectures, we compared them with quantum-inspired versions. The primary evaluation metric employed was the F2 score, chosen for its emphasis on penalizing false negatives more heavily than other metrics.

Our results demonstrate a distinct advantage of using quantum-inspired algorithms, particularly notable in the enhanced inference time. This improvement suggests that quantum-inspired methods can significantly accelerate the detection and prevention processes in AML applications, providing a practical benefit over traditional approaches.

Looking ahead, there are opportunities for future research that warrant exploration. One key area of interest is to examine how the observed benefits scale when the dimensions of the dataset are varied. This involves assessing whether the advantages in training and inference speed for neural networks and the accuracy improvements associated with QBoost are sustained as the dataset size increases.

Additionally, as the dataset expands, it may necessitate an increase in the number of weak learners within the QBoost algorithm. This potential increase could lead to a rise in the optimization complexity, posing challenges to the feasibility of the quantum-inspired approach. Investigating these aspects will be crucial to understanding the scalability and practical limitations of applying quantum-inspired algorithms to large-scale AML datasets.

Overall, our findings contribute to the growing body of evidence supporting the integration of quantum-inspired techniques in machine learning, particularly for complex tasks such as AML. By further exploring the scalability and optimization challenges, future research can build on our work to develop even more efficient and accurate AML detection systems.

\section*{Acknowledgments}
\label{sec:acknowledgments}
\vspace{-5pt}
This work is a collaboration between Bundesdrekerei Group's Innovation Hub and Multiverse Computing. We would like to acknowledge and thank the technical teams and field engineers both at Bundesdrekerei Group and Multiverse Computing for extremely helpful discussions.

\bibliographystyle{apsrev4-1} 
\bibliography{QCAPaper} 

\end{document}